\documentclass{article}

\usepackage[final]{neurips_2025}

\usepackage{amsmath}
\usepackage[utf8]{inputenc} % allow utf-8 input
\usepackage[T1]{fontenc}    % use 8-bit T1 fonts
\usepackage{url}            % simple URL typesetting
\usepackage{booktabs}       % professional-quality tables
\usepackage{amsfonts}       % blackboard math symbols
\usepackage{nicefrac}       % compact symbols for 1/2, etc.
\usepackage{microtype}      % microtypography
\usepackage{xcolor}         % colors
\usepackage{graphicx}

\def\eg{\emph{e.g.}} 
\def\vs{\emph{v.s.}} 
\def\ie{\emph{i.e.}} 
\def\cf{\emph{cf.}} 

\newcommand{\mytext}{CEPE$^*$}
\newcommand{\ourfullname}{\textsc{Vision Centric Token Compression}} %add by alex
\newcommand{\ourname}{\textsc{Vist}}
\newcommand{\our}{\textsc{Vist}}
\usepackage{multirow}
\usepackage{multicol}
\definecolor{bblue}{RGB}{0,30,95}
\definecolor{rred}{RGB}{190,0,0}
\definecolor{mygray}{gray}{.9}
\definecolor{myred}{RGB}{168,71,81}
\definecolor{ggray}{RGB}{127,127,127}
\definecolor{sblue}{RGB}{0,173,206}
\definecolor{ppink}{RGB}{240,46,142}
\definecolor{myblue}{RGB}{206,219,242}
\usepackage[table]{xcolor}
\usepackage{bm}
\usepackage{natbib}
\usepackage{bbding}
\usepackage{pifont}
\newcommand{\cmark}{\ding{51}}
\usepackage{booktabs}
\definecolor{darkergreen}{RGB}{21, 152, 56}
\definecolor{red2}{RGB}{252, 54, 65}

\newcommand\greenp[1]{\textcolor{darkergreen}{(#1)}}
\usepackage[toc]{appendix} % 用于生成附录目录
\usepackage{tikz} 

\newcommand{\myred}[1]{\textbf{\textcolor{myred}{#1}}}
\usepackage[normalem]{ulem} % 用于删除线效果
\usepackage{pifont}
\usepackage{caption}
\usepackage{wrapfig}
\usepackage{subcaption}
\usepackage{xcolor}
\definecolor{citecolor}{HTML}{0071BC}
\definecolor{linkcolor}{HTML}{ED1C24}
\usepackage[pagebackref=false, breaklinks=true,letterpaper=true, colorlinks,citecolor=citecolor, linkcolor=linkcolor, bookmarks=false]{hyperref}

\usepackage[textsize=tiny]{todonotes}
\newcommand{\colorboxinline}[1]{\textcolor{myblue}{\rule{0.8em}{0.8em}}}
\usepackage[utf8]{inputenc}
\usepackage{newunicodechar}
\newunicodechar{♫}{\ding{14}} % or another suitable symbol

\title{Vision-centric Token Compression \\in Large Language Model}

\author{
  Ling Xing$^{1*}$, Alex Jinpeng Wang$^{2*}$, Rui Yan$^{1\dagger}$, Xiangbo Shu$^{1}$, Jinhui Tang$^{3}$\\
  \small{$^1$Nanjing University of Science and Technology \quad $^2$Central South University \quad} \\
  \small{$^3$Nanjing Forestry University} \\
}

\begin{document}

\maketitle

\begin{abstract}
Real-world applications are stretching context windows to hundreds of thousand of tokens while Large Language Models (LLMs) swell from billions to trillions of parameters.
This dual expansion send compute and memory costs skyrocketing, making \emph{token compression} indispensable.
We introduce \ourfullname\ (\ourname), a \textit{slow–fast} compression framework that mirrors human reading: 
the \emph{fast} path renders distant tokens into images, letting a \textbf{frozen, lightweight vision encoder} skim the low-salience context; 
the \emph{slow} path feeds the proximal window into the LLM for fine-grained reasoning.
A Probability-informed Visual Enhancement (PVE) objective masks high-frequency tokens during training, steering the Resampler to concentrate on semantically rich regions—just as skilled reader gloss over function words.
On eleven in-context learning benchmarks, \ourname\ achieves the same accuracy with 2.3$\times$ fewer tokens, cutting FLOPs by 16\% and memory by 50\%.
This method delivers remarkable results, outperforming the strongest text encoder-based compression method CEPE by \textbf{7.6}\% on average over benchmarks like $_{\!}$TriviaQA, $_{\!}$NQ, $_{\!}$PopQA, $_{\!}$NLUI, and $_{\!}$CLIN, $_{\!}$setting a new standard for token efficiency in LLMs.
The project is at \url{https://github.com/CSU-JPG/VIST}.
\end{abstract}

\renewcommand{\thefootnote}{}
\footnotetext[1]{*\hspace{0.2em}Equal contribution}
\footnotetext[2]{\dag\hspace{0.2em}Corresponding author}
\renewcommand{\thefootnote}{\svthefootnote}
\section{Introduction}
\label{sec:intro}
Large language models (LLMs) excel at short snippets, yet many real-world tasks, \eg, long-document understanding~\cite{brown2020language,qwen} and question answering~\cite{sugawara2020constructing,trivedi2022musique}—already require inputs far beyond the thousand-token regimes of early GPT-3~\cite{brown2020language}.
At the same time, parameter counts have leapt from billions to trillions~\cite{achiam2023gpt,yin2025pangu,li202452b}.
In this dual squeeze of \emph{longer context \& larger models}, \textbf{compression shifts from a convenience to a necessity}: without shrinking the input, even the most powerful LLM cannot afford to reason over the information we want it to see.

\begin{figure}[ht]
    \centering
    \begin{subfigure}[t]{0.49\textwidth}
        \centering  \includegraphics[width=\linewidth]{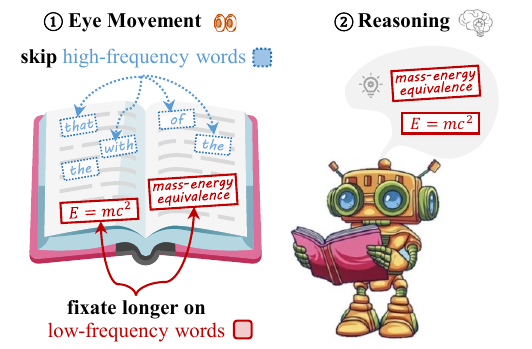} 
        \caption{\centering
        Selective reading strategy based on word frequency in skilled readers.}
        \label{fig:sub1}
    \end{subfigure}
    \hfill 
    \begin{subfigure}[t]{0.49\textwidth}
        \centering
\includegraphics[width=\linewidth]{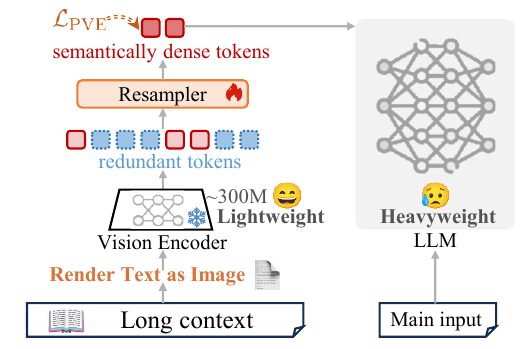}
        \caption{ \centering
        A visual pathway to unlock LLM context.}
        \label{fig:sub2}
    \end{subfigure}
    \caption{
    \small{
    Our method \textbf{\ourname\ }adopts a lightweight vision encoder to process loosely relevant long contexts, offering a \textbf{more cost-efficient alternative} to full LLM processing. 
    However, the inherent redundancy in long text leads to redundant visual tokens. 
    Motivated by \textbf{Selective Reading Strategy} where \textit{low-frequency (content) words receive longer fixations while high-frequency function words are often skipped}, we design Probability-informed Visual Enhancement (\ie, \(\textcolor[RGB]{197,90,17}{\mathcal{L}_{\text{PVE}}}\)). 
    This guides the Resampler to prioritize informative content over redundancy, \textbf{resulting in a 75\% reduction in the number of visual tokens} and yielding semantically dense tokens.
        }
    }
    \label{fig:motivation}
\end{figure}

Psycholinguistics shows that our eyes dance across text: we \emph{fixate} on rare, content-rich words and \emph{skip} almost one-third of high-frequency function words~\cite{rayner2006effect,strukelj2018one,gu2024skimmers}.
This \emph{selective-reading} strategy forms a natural \emph{slow-fast} circuit. 
A \emph{fast visual pass} skims distant, low-salience context to maintain global context, while a \emph{slow cognitive pass} focuses on nearby sentences that matter.
(Figure$_{\!}$~\ref{fig:motivation} (a)).

Motivated by this circuit, we present \ourfullname\ (\textbf{\ourname}), a \textit{slow-fast} token compression framework that mirrors human skimming.
As illustrated in Figure~\ref{fig:motivation} (b), \ourname\ first converts loosely relevant long context into images, which are processed by a frozen vision encoder and a trainable Resampler to produce semantically compact visual tokens. 
These compressed tokens and the main input tokens are then consumed by the LLM.
In this \textit{slow-fast} setup, the \textit{vision encoder acts like the human eye}—selectively attending to salient information—while \textit{the LLM functions as the brain}, concentrating on the most informative content for deeper reasoning.

Specifically, the frozen visual encoders (\eg, CLIP~\cite{radford2021learning}) trained on paired image-text data naturally \emph{acquire OCR capabilities}$_{\!}$~\cite{radford2021learning, lin2025parrot}, making them a powerful tool for image-based text understanding.
However, the inherent redundancy in long text leads to redundant visual tokens. 
To address the problem, we design \textbf{Probability-informed Visual Enhancement} (PVE), a contrastive scheme that enforces Resampler to prioritize informative content over redundancy.
Concretely, PVE applies \textbf{frequency-based masking strategy} to text token embeddings from the LLM tokenizer, suppressing high-frequency (less informative) text tokens. 
This semantically rich text supervision guides the Resampler to focus on informative content, bridging the semantic gap between visual and text tokens, and enabling more effective token compression.
Unlike previous work~\cite{li2023compressing,jiang2023llmlingua,nottingham2024selective,jiang2023longllmlingua,pan2024llmlingua} that rely on LLMs to compute token-level information entropy$_{\!}$ for$_{\!}$ assessing$_{\!}$ importance,$_{\!}$ \ourname\ adopts$_{\!}$ token$_{\!}$ frequency$_{\!}$ as$_{\!}$ a$_{\!}$ simple$_{\!}$ yet$_{\!}$ effective$_{\!}$ proxy,$_{\!}$ and$_{\!}$ further \textbf{reveals rare tokens are key contributors $_{\!}$to $_{\!}$overall $_{\!}$semantic meaning} (\cf~$_{\!}$Figure~$_{\!}$\ref{fig:info_gain} $_{\!}$and$_{\!}$ \S\ref{sec::discussion}).

\ourname\ leverages a lightweight vision encoder to compress loosely relevant long contexts, offering a cost-efficient alternative to full-scale LLM computation.
Furthermore, the vision encoder serves as a \textit{visual text tokenizer}, offering several compelling advantages over traditional text tokenizers.
\ding{182} \textbf{Simplified Tokenization.} Text tokenizers rely on complex tokenization rules and vocabulary constraints, typically involving nearly ten human-defined preprocessing steps (\eg, lowercasing, punctuation and stop word removal, and tokenization)~\cite{gage1994new}. 
However, the vision encoder processes text more directly by treating rendered text images as visual inputs. 
\ding{183} \textbf{Vocabulary Bottleneck Mitigation.} Text tokenization, constrained by a finite vocabulary, becomes a bottleneck when scaling to many languages. A larger vocabulary increases memory and computational costs in the embedding matrix and output layer.
However, vision encoder eliminates the need for text tokenizers and unifies various languages into a single image format that removes the need for a vocabulary~\cite{tai2024pixar, rust2022language}.
\ding{184} \textbf{Robustness to Character-Level Noise.} Vision encoders are more resilient to typos and low-level orthographic attacks, as they capture holistic visual patterns rather than relying on discrete token matching~\cite{rust2022language}.
\ding{185} \textbf{Multilingual Efficiency.} While our work focuses on English, visual text tokenizer can reduce the number of tokens compared to traditional text tokenizer for languages (\eg, 62\% for Japanese, 78\% for Korean, and 27\% for Chinese). This reduction is particularly impactful in long-text scenarios.
Taken together, \textit{leveraging vision encoders for long-context compression is a promising and worthwhile direction to explore}.

To validate the effectiveness of \ourname, we primarily compare with the text-encoder-based token compression counterpart CEPE~\cite{yen2024long}.
\ourname\ requires \textbf{2.3$\times$ fewer visual tokens} than text tokens for the same input, reducing \textbf{FLOPs by 16\%} and \textbf{memory usage by 50\%}.
\ourname\ also delivers consistent gains over CEPE on both In-Context Learning and Open-domain Question Answering tasks, with \textbf{average gains of 3.6\% across 11 datasets} and \textbf{5.7\% across 3 datasets}, respectively—highlighting the effectiveness of visual representations for long-context modeling in LLMs.
\section{Related Work}
\label{sec:related}

% \alex{Memorying Transformer, CEPE etc.}
% \todo{too long, cut down to 80\%}
% \alex{\subsection{Token Compression} }
\noindent\textbf{Token Compression.} 
There has been a growing
interest in expanding the context window for LLMs. 
A line of methods leverages LLM itself to compress raw long input. 
One may classify these works into two principal groups. \textbf{i)} \textit{soft prompt-based}  methods that adapt LLMs to compress context into fewer tokens~\cite{wingate2022prompt,zhang2024soaring,mu2024learning,chevalier2023adapting,gecontext}.
\textbf{ii)} \textit{selection-based} methods that remove redundant tokens based on information entropy computed by LLMs~\cite{li2023compressing,jiang2023llmlingua,nottingham2024selective,jiang2023longllmlingua,pan2024llmlingua,li2024snapkv}. 
All the long inputs typically need to be handled by the heavy LLMs, which incur high costs. 
Another line of work~\cite{mohtashami2023random,tworkowski2024focused} augments LLMs with the capacity to memorize previous long context information by external memory bank and retrieve relevant knowledge~\cite{xuretrieval,zhang2023retrieve,wumemorizing,wang2024augmenting}.
Our method is orthogonal to these existing strategies and can be combined with them to achieve longer context length. 
The most related work is CEPE~\cite{yen2024long}, which employs a lightweight text encoder to handle long contexts and integrates the information into LLM via cross-attention.
While CEPE reduces workload on the LLM, it overlooks the redundancy in long text, making it harder for LLMs to effectively allocate attention to key content. 
In contrast, \ourname\ compresses long text into compact visual tokens guided by high-density semantic text tokens.

% \alex{\subsection{Vision-centric Method}}
% \alex{PIXEL, CLIPPO etc}
\noindent\textbf{Vision-centric Method.}
% \todo{Do not exceed 10 lines for each paragraph. Pls check all paper.}
Text tokenization~\cite{kenton2019bert,kudo-richardson-2018-sentencepiece, sennrich-etal-2016-neural} breaks down text into tokens, serving as a fundamental step in natural language processing. 
However, tokenization-based methods lack robustness against spelling errors and face vocabulary bottlenecks. 
A new line of work tackles these issues in a tokenizer-free paradigm~\cite{salesky2021robust, rust2022language, gao2024improving}. 
The representative method Pixel~\cite{rust2022language} renders text as images and learns to reconstruct masked image patches at the pixel level. 
It demonstrates strong cross-language translation capabilities and tolerance for text perturbation. 
Along this direction, recent work explores different pre-training objectives~\cite{chai2024dual,lotz2023text}, \eg, contrastive learning~\cite{xiao2024pixel}, patch-and-text prediction~\cite{gao2024improving}.
Despite advancements, these methods overlook long-text scenarios and rely on complicated training pipelines, \eg, OCR-based text understanding~\cite{tai2024pixar}. 
In contrast, \ourname\ directly processes text images by leveraging a vision encoder pretrained on image-text pairs with strong OCR capabilities, and enhancing visual features using enriched text embeddings from LLM tokenizer.
An emerging family of multimodal methods~\cite{kim2022ocr, lee2023pix2struct, tschannen2023clippo,wang2024leveraging,li2023text} leverage visual representations to process text and images together, enabling a wide range of applications involving visually-situated text, \eg, webpage parsing~\cite{lee2023pix2struct}, tables images analysis~\cite{zhang2023mpmqa}, and document understanding~\cite{kim2022ocr, hu2024mplug}. 
In this work, we explore incorporating long-context information into LLMs from a visual perspective.

\section{Methodology}
\label{sec::method}
\begin{figure}[!t]
    \centering
    \includegraphics[width=0.95\textwidth]{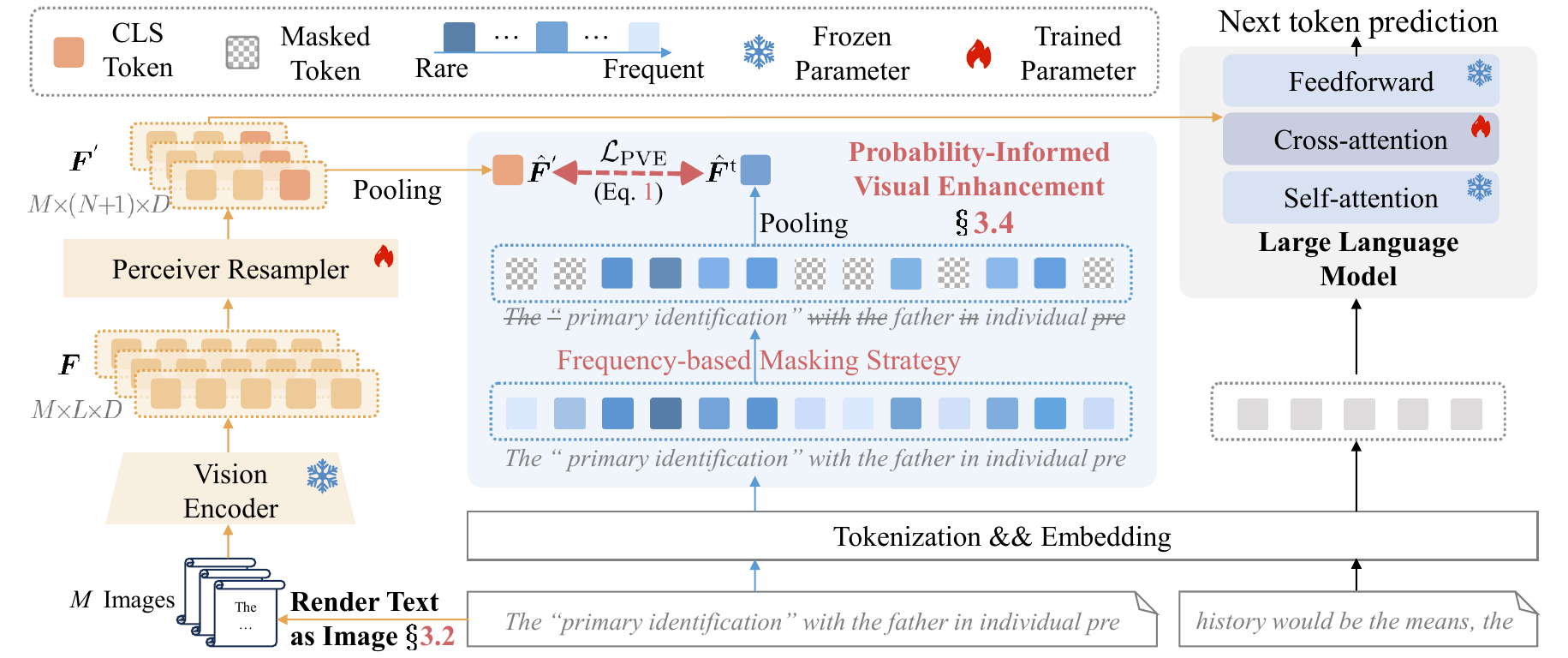} 
    \small
  \vspace{-3pt}
    
    \caption{\small{$_{\!}$\textbf{Overview of \textsc{Vist}}. 
    \textsc{Vist}, a \textit{slow-fast} token compression framework, efficiently processes long texts by mimicking human skimming. 
    First, the \textit{fast visual path} converts long context into images and employs a lightweight vision encoder to capture semantically compact visual features. 
    These features are then integrated into the LLM via cross-attention in the \textit{slow cognitive path}, allowing LLM to focus on salient content for deeper reasoning.
    To prioritize informative content in text images, \textsc{Vist} employs \textbf{Frequency-based Masking} on text token embeddings from text tokenizer, suppressing high-frequency but low-information token  \colorboxinline{blue} (\eg, ``the'' and ``with'').
    Such refined embeddings guide the Resampler in extracting critical semantics from the images. 
     }
     }
    \label{fig:overview}
\vspace{-7pt}
\end{figure}

% \alex{Sub Section1: Overall Pipeline.}

% \alex{Sub Section2: Vision-centric Implementation}

% \alex{Sub Section3: Token Reduction }
% \ourname\ extends the context window of existing Large language models (LLMs) by visual tokens and handles long input more efficiently. 
In this section, we present our method \ourname, which processes long in-context text by a lightweight visual encoder, effectively and efficiently extending the context length of LLMs.

% First, we outline the overall pipeline of \ourname. 
% Next, we explain how long texts are rendered into RGB images (\S\ref{sec::vision-centric}) and then converted into compressed visual tokens using a frozen vision encoder and a trainable Perceiver Resampler~\cite{alayrac2022flamingo} (\S\ref{sec::token_reduction}). 
% Finally, we present Probability-Informed Visual Enhancement (\S\ref{sec::FCL}), which guides Perceiver Resampler to extract rich textual information from rendered text images by using refined text token embeddings as supervision signals. 

% To address the redundancy in long texts, we integrate a frequency-based masking strategy into the loss, enhancing the information density of the text token embeddings.

% \alex{To Do: Plot a overall pipeline, especially about token reduction and vision-centric implementation part}

\subsection{Overall Pipeline}

Our \ourname, a \emph{slow-fast} compression framework, is designed to efficiently process long texts by mimicking human reading.
The \emph{fast visual path} skims distant, low-salience long context via a lightweight vision encoder, while the \emph{slow cognitive path} performs fine-grained reasoning on important content by LLM.
As illustrated in Figure~\ref{fig:overview}, the input long text (\ie, $T$ text tokens) is split into two parts: the first $T_\text{e}$ text tokens processed in a visual view and the remaining $T_\text{d}$ raw text tokens given to LLM, where $T = T_\text{e} + T_\text{d}$.  
Specifically, the $T_\text{e}$ text tokens are evenly rendered into $M$ images and fed into a frozen vision encoder. Then \ourname\ employs a learnable Perceiver Resampler to compress text-rendered image features into a fixed count of tokens. Such compressed visual tokens are integrated into the LLM via cross-attention for the next-token prediction. 
The Perceiver Resampler is jointly trained with the LLM during tuning the cross-attentions in an end-to-end manner.

To empower the model with the ability to comprehend dense text in images, we devise \textbf{Probability-informed Visual Enhancement} (PVE, \S\ref{sec::FCL}). 
PVE maximizes agreement between visual features obtained from the Perceiver Resampler and \textbf{text token embeddings extracted from LLM tokenizer}. 
This alignment bridges the global semantic gap between visual tokens and raw text tokens. 
Furthermore, to address token redundancy, \ourname\ incorporates a \textbf{frequency-based masking} mechanism within PVE that selectively masks high-frequency, low-information text tokens, thereby improving the information density of the text embeddings. 
These refined embeddings serve as enriched supervision signals, encouraging visual features to be more compact and semantically meaningful.

% This additional mechanism strengthens the synergy between visual and textual modalities, ultimately improving the model's capacity to process and interpret text in visual contexts.

% long text often contains redundant or less important information, meaning that not all text tokens contribute equally to semantic understanding.

\subsection{Vision-centric Implementation}
\label{sec::vision-centric}
\ourname\ transforms raw textual data into $M$ uniformly distributed RGB images $\mathcal{X}\!=\!\{x_m \!\in\! \mathbb{R}^{H\times W \times C}\}_{m=1}^{M}$, where $M$ can be dynamically adjusted based on the length of the input text. Concretely, each image is configured with height $H=14$, width $W=3,584$, and $C=3$ RGB channels, which corresponds to a square color image with a resolution of $224\times224$. 
Text is rendered using a 10px font size and Google \textit{Noto Sans} typeface. 
% On average, one Llama token requires approximately 1.28 $14\times14$ image patches. 
If text incompletely fill the image, white empty patches are masked to exclude them from attention score computation and loss calculation. 
Compared to text tokenizer-based methods, this rendering method does not lead to slower training speeds~\cite{tschannen2023clippo}.
                                        
\subsection{Token Reduction}
\label{sec::token_reduction}
The $M$ text-rendered images are first processed by frozen vision encoder, specifically the ViT-L/14$_{\!}$~\cite{radford2021learning} from OpenCLIP. 
The extracted features $\bm{F} \!\in\! {\mathbb{R}}^{M \times L \times D}$ are then fed into a trainable Perceiver Resampler$_{\!}$~\cite{alayrac2022flamingo}, producing a fixed set of $N\!+\!1$ visual tokens per image (including a CLS token), denoted as $\bm{F}^{'} \!\in\! {\mathbb{R}}^{M \times (N+1) \times D}$, where $N=64$ and $D$ is the feature dimension.
During training, raw text data ($T_\text{e}=4096$ text tokens) is rendered onto $M=28$ images, resulting in $64 \times 28 = 1792$ visual tokens, passed to the cross-attention layer in LLM. 
This compression reduces the computational complexity of the cross-attention layer within the LLM. 
Moreover, the number of images $M$ and tokens $N$ can be dynamically adjusted during both training and inference, \textit{allowing \ourname\ to flexibly control the compression ratio}. 
\ourname\ using a lightweight vision encoder, offers a more efficient approach than processing all text tokens directly within the LLM. 
% extract rich text information from pixel space
\subsection{Probability-informed Visual Enhancement}
\label{sec::FCL}
In \ourname, the frozen vision encoder is pre-trained primarily on general visual data (such as natural images) without exposure to rendered text images. Hence its ability to interpret dense textual information within images is constrained. 
To alleviate this problem, we develop a novel training objective, named Probability-informed Visual Enhancement (PVE). 
PVE enhances the understanding capabilities of Perceiver Resampler for rendered text images, enabling them to serve as robust substitutes for traditional text tokenizers.
% by leveraging text token embeddings extracted from tokenizer in LLM. 

% FCL leverages text token embeddings to guide the resampler in extracting textual information from text images. However, when dealing with long texts, these embeddings often contain redundant information, which hampers the resampler's ability to capture meaningful semantic information. To mitigate this, we integrate a frequency-based masking strategy into FCL, which randomly removes less informative tokens (\ie, high-frequency tokens) from the text token embeddings. This ensures the resampler focuses on crucial content-bearing tokens. With these improvements,  Perceiver Resampler in \ourname\ can learn richer semantic representations, leading to more effective interpretation of text within images. 

% enforces text token embeddings (extracted from text tokenizer in LLM) and visual text features (from Perceiver Resampler) to a joint semantic space, where text token embeddings serve as free supervision signals.

% aims to equip the vision encoder and Perceiver Resampler with the ability to interpret visual text in rendered images and serve as a robust substitute for traditional text tokenizers. FCL makes efforts to map text token embeddings (extracted from text tokenizer in LLM) and visual text features (from Perceiver Resampler) to a joint semantic space, relying on text token embeddings to serve as free supervision signals.
% (processed by frequency-based masking strategy) 
\begin{figure}[t]
    \centering   \includegraphics[width=0.98\textwidth]{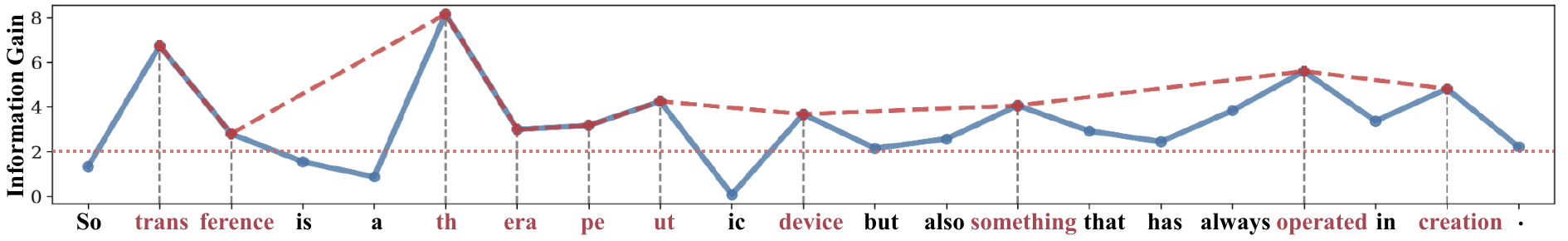} 
    \vspace{-3pt} 
   
\caption{
\small{
\textbf{Token-Level Information Gain} (IG) in sentence ``\textit{So \myred{transference} is a \myred{therapeut}ic \myred{device} but also \myred{something} that has always \myred{operated} in \myred{creation}.}''. The red dashed line masks of 50\% the most frequent tokens based on training set statistics. 
This strategy preserves tokens with higher information gain, while eliminating statistically prevalent but low-value tokens, enhancing semantic density.
}
}
 \label{fig:info_gain}
 \vspace{-10pt}   
\end{figure}

\noindent\textbf{Text-anchored Semantic Consistency.}
PVE encourages the Perceiver Resampler to learn a shared embedding space, aligning visual text features $\bm{F}^{'}$ with text token embeddings from text tokenizer. Concretely, 
PVE is formulated as a contrastive loss:
\begin{equation}
\label{eq:FCL}
    \mathcal{L}^{ij}_{\text{PVE}} = -\log \frac {\exp(\langle \hat{\bm{F}}^{'}_{i}, \hat{\bm{F}}^\text{t}_j \rangle / \tau)} 
    {\sum_{k=1}^B \exp (\langle \hat{\bm{F}}^{'}_{i}, \hat{\bm{F}}^\text{t}_k \rangle /\tau )},  
\end{equation}
where $B$ is batch size and $\hat{\bm{F}}^{'}_{i}$ is obtained by applying average pooling to the CLS tokens from $\bm{F}^{'}_{i}$.
$\hat{\bm{F}}^\text{t}_j$ is the averaged text token embedding after \textbf{frequency-based masking} and pooling.
$\tau$ is the temperate parameter. 
Importantly, $\hat{\bm{F}}^{'}_{i}$ and $\hat{\bm{F}}^\text{t}_i$ are different representations derived from the same text.

\noindent\textbf{Frequency-based Masking.} 
PVE employs text token embeddings as supervision signals to guide the Resampler in extracting textual information from text images. 
However, long-form text is inherently redundant, where structural components and function words may dominate the token distribution.
Such redundancy introduces noise that impedes Resampler from capturing \textbf{key semantic content}.
% However, text tokens often contain redundant information, especially in long texts, which hinders Resampler from capturing key semantic content.
% Not all text tokens contribute equally to semantic understanding, but how can we \textit{quantify the amount of information contained in text}? 
% \input{tabs/ppl}

Our solution draws inspiration from Shannon information theory~\cite{shannon1948mathematical}, which provides a formal way to quantify the information content of an event or a message. 
The formula is given by:
\begin{equation}
I(y) = -\log_{2} P(y),
\end{equation} where $I(y)$ is the information content of event or messages $y$ and $P(y)$ is the probability of $y$. 
It highlights the inverse relationship between the probability of an event and the information it carries. 
When applied to tokens in a corpus: \textbf{Rare tokens} (low-frequency) are treated as high-information tokens because they often carry domain-specific or contextually important information. 
\textbf{Frequent tokens} (high-frequency) have lower information content because they may serve more structural or grammatical purposes, contributing less to the unique meaning of the text. 
Figure~\ref{fig:info_gain} shows that masking 50\% of the most frequent tokens based on corpus-level (\ie, training set) frequency distribution still \textit{preserves most high-information-gain (IG) tokens, ensuring minimal loss of critical information while reducing redundancy}.
This aligns with the \textbf{selective reading strategies}~\cite{gu2024skimmers,brysbaert1998word} observed in skilled readers.
Based on this principle, we devise frequency-based masking strategy that uses token frequency as a proxy for semantic importance.  
This strategy masks frequent tokens but low-information tokens to improve the information density of text token embeddings. 
The importance score for each token is calculated as follows: 
\vspace{-3pt}
\begin{equation}
s_w = \log \frac{|\mathcal{S}|}{1+\text{count}(w)},
\label{eq:important_score}
\end{equation}
where $|\mathcal{S}|$ denotes the total number of samples, $\text{count}(w)$ is the count of the token $w$ (subword), and $s_w$ is the importance score of token $w$. 
Token frequency statistics can be easily computed online with negligible overhead or precomputed.
Based on the importance score for each token, we apply a 50\% masking rate, where \textit{tokens are randomly masked with tokens of lower importance score being more likely to be masked}. 
This ensures the Resampler prioritizes key content-bearing tokens, learning richer semantic representations and improving its ability to interpret dense text in rendered images.

\section{Experiment}
\label{sec::experi}
\subsection{Experimental Setup}
\noindent\textbf{Pretraining.}$_{\!}$ We validate \ourname\ with TinyLlama~\cite{zhang2024tinyllama}. The frozen vision encoder in our model is ViT-L/14~\cite{radford2021learning}. 
To reduce computational overhead, our model employs float16 precision and DeepSpeed Zero-2 with CPU off-loading~$_{\!\!}$\cite{rasley2020deepspeed}.  
% Refer to Appendix~\ref{secb} for details.
% 文本的一些处理：增强。。  在方法部分写？
% \noindent\textbf{Visual text.} The input of vision encoder, \ie, $4,096$ text token sequences, 

\noindent\textbf{Competitors.}
\textbf{i) Long-context models:} Replug$_{\!}$~\cite{shi2024replug} and Stream$_{\!}$~\cite{xiaoefficient} with TinyLlama$_{\!}$~\cite{zhang2024tinyllama}.
\textbf{ii) Text-encoder-based compression method:} To compare the effectiveness of leveraging text tokens \vs visual tokens for processing long contexts in LLM, we implement \mytext\ by applying CEPE~\cite{yen2024long} to TinyLlama~\cite{zhang2024tinyllama}, replacing the vision encoder in \ourname\ with a lightweight text encoder.
All other architectural and training settings are kept identical.
\textbf{iii) Vision-centric compression methods:}  
ToMe~\cite{bolyatoken} merges, and FastV~\cite{chen2024image} prunes visual features from frozen vision encoders. 
For fairness, we match their compression rates to ours.
Directly using these visual features in LLMs leads to high perplexity (>1k) due to the mismatch between visual and text tokens.
Thus, we retain the Resampler and PVE in our \ourname, denoting ToMe$^\dagger$ and FastV$^\dagger$.
% See Appendix~\ref{appendix:vision_centric_details} for more details.

% It is worth noting that, the difference between \mytext\ and our method lies in how in-context texts are processed.  Such text encoder follows the configuration of RoBERTa-large~\cite{liu1907roberta}, as in CEPE~\cite{yen2024long}.
 
\noindent\textbf{Pretraining Dataset.} Our pertaining dataset is an official sample of the RedPajama dataset~\cite{weberredpajama}, including $1$B tokens from seven domains: ArXiv, Book, C4, Commoncrawl, GitHub, StackExchange, and Wikipedia. The training set of the corpus is preprocessed into $4608$ text token sequences, where the first $4096$ text token sequences are fed into the vision encoder (or text encoder for \mytext) and the remaining $512$ text tokens are provided to LLM.

\noindent\textbf{Downstream Evaluation.} 
We primarily evaluate tasks requiring long context processing, revealing vision tokens effectively handle extended context, outperforming previous text-encoder-based model. 
% Comparisons across more methods and datasets are provided in the Appendix~\ref{appendix::vision-centric-method}.

% \input{tabs/ppl}
\begin{table}[t] % 使用 table* 环境支持跨双栏
% \vspace{-10pt}

\caption{\small{Text perplexity on the last 256 tokens of long-context language modeling for ArXiv and Book datasets, PG19, Proof, and Code.
$T_\text{e}$ is token length for encoder, and $T_\text{d}$ is for LLM. $\dagger$ denotes methods with the Resampler and PVE in our \ourname. We report the Throughput of each model relative to TinyLlama. 
$\Delta$ is compression ratio.}
}
\vspace{-4pt}
\label{tab:ppl}
% \vskip 0.15in
\begin{center}
% \begin{small}
 % 动态调整图形和标题之间的间距
  % \ifdim\baselineskip>0.1in
  %   \vspace*{-\baselineskip} % 清除默认间距
  %   \vspace{0.05in}           % 设置固定间距为 0.1 英寸
% \setlength\tabcolsep{9pt}
\renewcommand\arraystretch{1.1}
\resizebox{1.0\textwidth}{!}{
\begin{tabular}{lccccccccccc}
    \toprule
        % &  &   & &&&& &
        % & \textbf{Compression} &\textbf{FLOPs}   
        % &\textbf{MEM}   \\
        \textbf{Method} & $T_\text{e}$ & $T_\text{d}$ & \textbf{ArXiv} & \textbf{Book} & \textbf{PG19} &\textbf{Proof}  & \textbf{Code}& \textbf{Throughput}& $\Delta$ &\textbf{TFLOPs} &\textbf{MEM}(GB) \\
        % \multicolumn{1}{c}{\multirow{-2}{*}{\hspace{0.5mm} \textbf{Method}   \hspace{0.5mm}}}
        % & \multicolumn{1}{c}{\multirow{-2}{*}{\hspace{0.5mm} $T_\text{e}$   \hspace{0.5mm}}}
        % & \multicolumn{1}{c}{\multirow{-2}{*}{\hspace{0.5mm} $T_\text{d}$   \hspace{0.5mm}}} 
        
        % &  
        % \multicolumn{1}{c}{\multirow{-2}{*}{\hspace{0.5mm}\textbf{ArXiv}   \hspace{0.5mm}}} 
        % &  \multicolumn{1}{c}{\multirow{-2}{*}{\hspace{0.5mm}\textbf{Books}   \hspace{0.5mm}}}            &  \multicolumn{1}{c}{\multirow{-2}{*}{\hspace{0.5mm}\textbf{PG19}   \hspace{0.5mm}}}     & \multicolumn{1}{c}{\multirow{-2}{*}{\hspace{0.5mm}\textbf{Proof}   \hspace{0.5mm}}}    & \multicolumn{1}{c}{\multirow{-2}{*}{\hspace{0.5mm}\textbf{Code}   \hspace{0.5mm}}}  &   \multicolumn{1}{c}{\multirow{-2}{*}{\hspace{0.5mm}\textbf{Latency}   \hspace{0.5mm}}}
        % &
        % \textbf{Ratio} 
        % &
        %  \textbf{(TFLOPs)}   
        % &
        %  \textbf{(GB)}  
        
        % \\
    
    \midrule

         % {\textbf{Total Tokens $= 2048$}}                             &                &  \\               
    % \midrule

    %     TinyLlama & - &   2048 & \textbf{2.978}         & 15.392 &           11.839 &  \textbf{2.824}  &    \textbf{2.189} & - & 4.24 &	4.78       \\
        
    %     \mytext & 1024 &   1024 &    3.259      & \textbf{14.558}  &    \textbf{11.515}       &           3.095        &  2.284 &    - & 3.75\greenp{0.49$\downarrow$} &	3.71\greenp{1.07$\downarrow$}    \\

    %      \our  & 1024 &   1024 &   3.243      & 14.983  &   13.987         &     3.588 &        2.600       &   2.3 & \textbf{2.94\greenp{1.30$\downarrow$}}	& \textbf{3.69\greenp{1.09$\downarrow$}}        \\
        
    % \midrule
        
    %     {\textbf{Total Tokens $= 4096$}}                             &                &               &               \\
    % \midrule
        
        TinyLlama$_{\!}$~\cite{zhang2024tinyllama}  & - &   4096& $>10^3$ & $>10^3$ & $>10^3$  &  $>10^3$  &  $>10^3$ & 1.0$\times$   & - & 8.47 & 5.46      \\
        
        Replug$_{\!}$~\cite{shi2024replug}  & - &  4096 & 3.220  & 15.394 & 14.685  & 3.921  & 3.011 & 0.2$\times$ & - & 9.15 &	6.12    \\
        
        Stream$_{\!}$~\cite{xiaoefficient} & - & 4096 &  3.116 & 15.188 &  14.372 & \textbf{2.876} & 2.764 & 1.6$\times$ & -& 8.31 &	6.41    \\

        ToMe$^\dagger$$_{\!}$~\cite{bolyatoken}  & 2048 &  2048 &  3.536 & 15.607 & 16.213  &  4.128 & 3.234 & 3.8$\times$ & 2.3 & 7.93 &	 4.59   \\
        FastV$^\dagger$$_{\!}$~\cite{chen2024image}  & 2048 &  2048 & 3.491  & 15.711 &  16.016 & 4.216  & 3.111 & 3.8$\times$ & 2.3 & 7.88 &	  4.59  \\
        
         \mytext$_{\!}$~\cite{yen2024long}  & 2048 &   2048& 3.071 &	15.619 &	\textbf{11.737} &	2.888 &	\textbf{2.151} & 1.8$\times$ & -   &8.26
         &	4.80 \\
         
        \our & 2048 &   2048 & \textbf{2.993}   & \textbf{14.973} &    13.205        &    3.057	&  2.247   & 3.8$\times$ & 2.3  & \textbf{7.72\greenp{0.75$\downarrow$}} &	\textbf{4.59\greenp{0.87$\downarrow$}}\\

    \midrule
    %     \textbf{Total Tokens $=  8192$}         \\
    % \midrule

        \mytext$_{\!}$~\cite{yen2024long} & 6144 &  2048& 3.005 & 14.919 &	\textbf{11.112} &	\textbf{2.719}  &	 \textbf{2.100} & 2.1$\times$& -  & 13.27&	7.74          \\
         
        \our & 6144 &  2048&   \textbf{2.989}   & \textbf{14.894}  &   12.737    &    3.003    & 2.183    &     \textbf{5.3$\times$} &   2.3  &  \textbf{11.65\greenp{1.62$\downarrow$}}	& \textbf{4.94\greenp{2.80$\downarrow$}}         \\
       
    \midrule
        
    %     \textbf{Total Tokens $= 16,384$}         \\
    % \midrule
        \mytext$_{\!}$~\cite{yen2024long} &  14,336  &     2048    &  3.003        &      14.921          &     \textbf{10.909} &     \textbf{2.715 }          &             2.162  & 3.3$\times$ & -   &23.30 &	13.59                     \\
         
        \our &   14,336 &     2048          &   \textbf{2.965}     &     \textbf{14.815}           &   11.933           &      2.971 & \textbf{2.032}    & \textbf{7.6$\times$} & 2.3  & \textbf{19.52\greenp{3.78$\downarrow$}}	& \textbf{6.75 \greenp{6.84$\downarrow$}}       \\
      
    \bottomrule
    \end{tabular}
}
% \end{small}
\end{center}
% \vskip -0.1in
\vspace{-15pt}
\end{table}

% Note use of \abovespace and \belowspace to get reasonable spacing
% above and below tabular lines.

 % &  &   & &&&& &
        % & \textbf{Compression} &\textbf{FLOPs}   
        % &\textbf{MEM}   \\
        % \textbf{Method} & $T_\text{e}$ & $T_\text{d}$ & \textbf{ArXiv} & \textbf{Books} & \textbf{PG19} &\textbf{Proof}  & \textbf{Code}& \textbf{Throughput}& \Delta &\textbf{TFLOPs} &\textbf{MEM}(GB) & \\
        % \multicolumn{1}{c}{\multirow{-2}{*}{\hspace{0.5mm} \textbf{Method}   \hspace{0.5mm}}}
        % & \multicolumn{1}{c}{\multirow{-2}{*}{\hspace{0.5mm} $T_\text{e}$   \hspace{0.5mm}}}
        % & \multicolumn{1}{c}{\multirow{-2}{*}{\hspace{0.5mm} $T_\text{d}$   \hspace{0.5mm}}} 
        
        % &  
        % \multicolumn{1}{c}{\multirow{-2}{*}{\hspace{0.5mm}\textbf{ArXiv}   \hspace{0.5mm}}} 
        % &  \multicolumn{1}{c}{\multirow{-2}{*}{\hspace{0.5mm}\textbf{Books}   \hspace{0.5mm}}}            &  \multicolumn{1}{c}{\multirow{-2}{*}{\hspace{0.5mm}\textbf{PG19}   \hspace{0.5mm}}}     & \multicolumn{1}{c}{\multirow{-2}{*}{\hspace{0.5mm}\textbf{Proof}   \hspace{0.5mm}}}    & \multicolumn{1}{c}{\multirow{-2}{*}{\hspace{0.5mm}\textbf{Code}   \hspace{0.5mm}}}  &   \multicolumn{1}{c}{\multirow{-2}{*}{\hspace{0.5mm}\textbf{Latency}   \hspace{0.5mm}}}
        % &
        % \textbf{Ratio} 
        % &
        %  \textbf{(TFLOPs)}   
        % &
        %  \textbf{(GB)}  
        
        % \\
    
\subsection{Long-context Language Modeling} 
To assess the long-context language modeling (LCM) ability, we evaluate on ArXiv and Book from RedPajama~\cite{weberredpajama} test split, alongside long-context datasets: PG19~\cite{raecompressive}, Proof~\cite{azerbayev2023proofpile}, and Code~\cite{codep}. 
The evaluation metric is perplexity (PPL) over the last 256 tokens of each input.
Early tokens (typically farther from the current prediction point) are handled by the vision encoder, while the more recent tokens are passed to the LLM, under the assumption that proximity correlates with relevance.

\noindent\textbf{Impact of Increased Text Length.}
Table~\ref{tab:ppl} summarizes the results across different input lengths.
Long-context language modeling can benefit from previous long contextual information. 
However, TinyLlama~\cite{zhang2024tinyllama} supports only fixed-size inputs of 2048 tokens. Beyond this length, its performance drops sharply, with perplexity exceeding $10^3$. 
In contrast, \ourname\ demonstrates a consistent decrease in perplexity as the input text length increases. 
% For instance, increasing the encoder input length from 2048 to 6144 reduces the perplexity on PG19 from 13.205 to 12.737. 
Vision-centric token compression models ToMe$^\dagger$ and FastV$^\dagger$ focus on natural images, where redundancy arises from local visual similarity.
However, they are \textit{ill-suited for text redundancy and struggle to preserve key semantic content in text image}, yielding higher perplexity than our \ourname.
Moreover, \ourname\ obtains the lowest PPL (14.973) on Book datasets, when $T_\text{e}$ and $T_\text{d}$ are 2048. 
These results prove that \textbf{ \ourname\ effectively enhances the capability of modeling long-form language.}

\noindent\textbf{Comparison on Inference Cost.} 
In Table~\ref{tab:ppl}, \ourname\ renders text into multiple images of size $224\times224$. 
1024 text tokens need 7 images and \ourname\ requires 56\% fewer visual tokens than text tokens for the same input (\ie, compression ratio $\Delta$ is 2.3, from 1024 text tokens to $448=7\times64$ visual tokens). 
We report the throughput of each model relative to TinyLlama.
\ourname\ achieves comparable performance with text-encoder-based compression model \mytext, with 16\% fewer FLOPs, 50\% less memory usage, and higher throughput when processing 16k tokens.

% Note that, compared to the text tokenizer in text encoder, a visual encoder requires only $43\%$ of the tokens to represent the same amount of text.

\begin{table}[t] 
% \vspace{-10pt}
\caption{\small{In-context learning accuracy averaged across 3 seeds (42, 43 and 44). 
Green highlights the gain from the additional demos. $n_\text{e}$ is the number of demos for encoder and $n_\text{d}$ for decoder (LLM).}
}
\label{tab:icl}
\vspace{-4pt}
\begin{center}

\renewcommand\arraystretch{1.2}
\resizebox{1.0\textwidth}{!}{
\begin{tabular}{lcccccccccccccc}
    \toprule
     \textbf{Method}  & $n_\text{e}$  & $n_\text{d}$  & \textbf{SST2} & \textbf{MR} & \textbf{AGN} & \textbf{SST5} & \textbf{NLUS} & \textbf{NLUI} & \textbf{TREC} & \textbf{TREF} & \textbf{DBP} & \textbf{BANK} & \textbf{CLIN} & \textbf{Avg.} \\
    \midrule
    % \textbf{Total Demonstrations $=2$} &   &  &   &   &   &   &   &   &   &   &  &  \\
% \midrule
    TinyLlama$_{\!}$~\cite{zhang2024tinyllama}  &  -& 2  & 76.0 & 67.7	 &63.4	 &27.6 &	5.2	 & 4.4 &	28.8	 &9.6 &	38.0 &	23.0	 &22.4 &	33.3 \\
\midrule

%     \textbf{Total Demonstrations $=20$}  &   &  &   &   &   &   &   &   &   &   &  & \\
% \midrule
    TinyLlama$_{\!}$~\cite{zhang2024tinyllama} & - & 20 &  \bf87.6  &	71.7  &	\bf75.0  &	30.1  &	\bf46.1  &	32.6  &	\bf72.0  &	\bf38.5  &	\bf80.4  &	\bf42.9  &	\bf53.7  & \bf57.3\greenp{24.0$\uparrow$} \\
    
    \mytext$_{\!}$~\cite{yen2024long}  &  18 &2 &  76.9 &	\bf82.3 &	 66.9 &	 29.1 &	 9.6 &	30 &	39.2 &	12.7 &	71.1 &	27.2 &	39.8 &	44.1\greenp{10.8$\uparrow$}  \\
    \our  &  18 &2 & 77.7 &	79.2 &	61.5 &	\bf42.7 &	15.6 &	\bf40.6 &	36.5 &	14.6 &	71.9 &	25.0 &	43.8 &	46.3\greenp{13.0$\uparrow$} \\
\midrule

    % \textbf{Total Demonstrations $=50$}  &   &  &   &   &   &   &   &   &   &   &  &  \\
    % \midrule
    TinyLlama$_{\!}$~\cite{zhang2024tinyllama}  & - & 50 & \bf88.6  & 64.8  & 21.4  & 42.5  &  \bf34.2 &  30.4 & \bf81.1  & \bf44.7  & 3.4  & \bf49.7  & 39.7 & 45.5\greenp{12.2$\uparrow$}  \\

    % Replug  & \\
    %  Stream & \\
    % ToMe$^\dagger$  & \\
    % FastV$^\dagger$  & \\
    
    \mytext$_{\!}$~\cite{yen2024long}  &  48 & 2 & 82.9  &	79.4 &	63.9  &	42.3  &	28.1 &	31.1 &	32.6 &	14.7 &	71.5 &	29.0 & 39.1	 &	46.8\greenp{13.5$\uparrow$}  \\
    \our  &  48 & 2 & 78.9  &	\bf85.2 &	\bf71.9  &	\bf44.4  &	27.2 &	\bf43.1 &	38.3  &	18.4 &	\bf73.1 &	25.4 &	\bf48.1 &	
    \bf 50.4\greenp{17.1$\uparrow$}  \\

    % \midrule
    % \llama$^\dagger$ & 40 & 94.3 & 98.7 & 74.7 & 52.3 & 87.7 & 54.8 & 95.1 & 76.7 & 62.1 & 50.4 & 72.0&  \\

    \bottomrule
    \end{tabular}
    }
% \end{small}
\end{center}
\vspace{-18pt}
\end{table}

% Note use of \abovespace and \belowspace to get reasonable spacing
% above and below tabular lines.
% \input{tabs/icl}
% \input{tabs/qa}

\subsection{In-context Learning}
We evaluate \ourname\ on In-Context Learning (ICL) tasks across 11 widely-used text-classification datasets: $_{\!}$SST2$_{\!}$~\cite{socher2013recursive}, MR$_{\!}$~\cite{pang2005seeing}, AGN$_{\!}$~\cite{zhang2015character}, SST5$_{\!}$~\cite{socher2013recursive}, TREC, TREF$_{\!}$~\cite{voorhees2000building}, 
 DBP$_{\!}$~\cite{zhang2015character}, NLUS, NLUI$_{\!}$~\cite{liu2021benchmarking}, 
BANK$_{\!}$~\cite{casanueva2020efficient}, and CLIN$_{\!}$~\cite{larson2019evaluation}. 
Following~\cite{yen2024long}, we randomly sample 250 text examples per dataset. 
The ICL results in Table~\ref{tab:icl} are reported as the average accuracy over three random seeds. For \ourname\ and \mytext, we provide two demonstrations directly to the decoder, while the rest are processed by the encoder. 

% Each demo is processed in parallel by text encoder in \mytext. Our method renders demos into a suitable number of images based on text length.

\noindent\textbf{Results.} 
Table~\ref{tab:icl} examines the influence of increasing the number of demonstrations, where $n_\text{e}$ is the number of demos for encoder and $n_\text{d}$ for LLM. 
\ourname\ shows a 13\% accuracy improvement (from 33.3\% to 46.3\%) as more demonstrations ($n_\text{e}$ is 18) are provided to the visual encoder, showcasing the capacity of LLM to comprehend text within visual signals when integrated with \ourname. 
Furthermore, \ourname\ outperforms \mytext\ in average accuracy across all 11 datasets, which indicates \textbf{visual-based text understanding can effectively match or even surpass text encoder performance}. 
Though \ourname\ and \mytext\ ($n_\text{e}=18,n_\text{d}=2$) underperform TinyLlama ($n_\text{d}=20$), they \textit{achieve lower computational cost by processing most demonstrations (18) with lightweight encoder}.
The performance gap on NLUS, TREC and TREF may stem from high category diversity, where the weak relevance between queries and demonstrations makes the lightweight encoding less effective than using the full LLM for all demos.
Notably, the performance of TinyLlama declines with 50 demonstrations due to context window limit, while \ourname\ remains efficient and stable.
\subsection{Open-domain Question Answering}
Open-domain Question Answering (QA) is a challenging task that requires model to generate accurate answers based on retrieved relevant information. 
Experiments are conducted on three open-domain QA datasets, including TriviaQA~\cite{joshi2017triviaqa}, NQ~\cite{kwiatkowski2019natural}, and PopQA~\cite{mallen2023not}. 
We use Contriever~\cite{izacardunsupervised} to retrieve relevant $k$ passages from Wikipedia, as in CEPE~\cite{yen2024long}. 
We prioritize passing the most relevant passages to the decoder to enhance performance. 
In Table~\ref{tab:qa}, we report the exact match (EM) scores.

% \input{tabs/qa}
% $k_\text{e}$ denotes the number of passages input to the encoder, while $k_\text{d}$ represents the number of passages processed by the decoder.

% \begin{wraptable}{r}{0.5\textwidth}
% % \vspace{-6pt}
% \caption{\textbf{Open-domain QA results.} $k_\text{e}$ represents the number of passages provided to the encoder, while $k_\text{d}$ denotes the number of passages given to the LLM. We report the exact match score.}
% \label{tab:qa}
% % \vspace{-12pt}

% \begin{center}
% \begin{small}
% \renewcommand\arraystretch{1.1}
% \resizebox{0.49\textwidth}{!}{
% \begin{tabular}{lcccccc}
%             \toprule
%            \textbf{Method} & $k_\text{e}$ & $k_\text{d}$ & \textbf{TriviaQA} & \textbf{NQ} & \textbf{PopQA} \\
%             \midrule
%             TinyLlama   & -  & 10  & 21.45  & 8.45 &	10.79 \\
%             TinyLlama   & -  & 15  & $<1$  & $<1$ &	$<1$ \\
%         \midrule
%             \mytext & 5 & 10 & 16.41	& 6.09	& 4.92 \\
%               \our  & 5 & 10 & \bf25.20\greenp{8.79$\uparrow$} &	\bf8.71\greenp{2.62$\uparrow$} & \bf11.44\greenp{6.52$\uparrow$}\\
%         \midrule
%               \mytext & 20 &10 & 16.56  & 6.75  & 5.78   \\
%               \our  & 20 &10 & \bf25.67\greenp{9.11$\uparrow$} & \bf8.81\greenp{2.06$\uparrow$}  & \bf11.84\greenp{6.06$\uparrow$}  \\

%             \bottomrule
%         \end{tabular}
% }

% \end{small}
% \end{center}
% % \vspace{-25pt}
% \end{wraptable}

\begin{wraptable}[14]{r}{0.5\textwidth}
\vspace{-12pt}
\centering
    \caption{\small{\textbf{Open-domain QA results.} $k_\text{e}$ represents the number of passages provided to the encoder, while $k_\text{d}$ denotes the number of passages given to the LLM. We report the exact match score.}}
    \vspace{-4pt}
    \renewcommand\arraystretch{1.5}
    \resizebox{\linewidth}{!}{ %< auto-adjusts font size to fill line
         \begin{tabular}{lcccccc}
            \toprule
           \textbf{Method} & $k_\text{e}$ & $k_\text{d}$ & \textbf{TriviaQA} & \textbf{NQ} & \textbf{PopQA} \\
            \midrule
            TinyLlama$_{\!}$~\cite{zhang2024tinyllama}   & -  & 10  & 21.45  & 8.45 &	10.79 \\
            TinyLlama$_{\!}$~\cite{zhang2024tinyllama}      & -  & 15  & $<1$  & $<1$ &	$<1$ \\
            \our  & 10 & 0 & 21.27 & 8.51 & 10.67\\
        \midrule
            \mytext$_{\!}$~\cite{yen2024long} & 5 & 10 & 16.41	& 6.09	& 4.92 \\
            
              \our  & 5 & 10 & \bf25.20\greenp{8.79$\uparrow$} &	\bf8.71\greenp{2.62$\uparrow$} & \bf11.44\greenp{6.52$\uparrow$}\\
        \midrule
              \mytext$_{\!}$~\cite{yen2024long} & 20 &10 & 16.56  & 6.75  & 5.78   \\
              \our  & 20 &10 & \bf25.67\greenp{9.11$\uparrow$} & \bf8.81\greenp{2.06$\uparrow$}  & \bf11.84\greenp{6.06$\uparrow$}  \\

            \bottomrule
        \end{tabular}
        }
    \label{tab:qa}
\end{wraptable} 
\noindent\textbf{Results.} 
TinyLlama is limited by a maximum context window of 2048 tokens, restricting it to processing no more than 10 passages at a time. 
Beyond this limit, performance drops sharply, with EM score falling below 1. 
For passages $k_\text{d} = 10$, the input already approaches or even exceeds the 2048-token limit of TinyLlama, so we truncate the input to avoid performance degradation.
When processing 5 extra passages (\ie, $k_\text{e}=5, k_\text{d}=10$), \ourname\ results in an EM score improvement of \textbf{3.75} compared to TinyLlama on TriviaQA~\cite{joshi2017triviaqa} dataset.
Moreover, it even surpasses text encoder-based approach \mytext\ under the same input conditions, \eg, delivering an EM score \textbf{9.11} points higher on the TriviaQA dataset when $k_\text{e}=20, k_\text{d}=10$. 
\textit{This enhancement may be attributed to PVE in \ourname\, which leverages enriched text embeddings to guide the Resampler in capturing key semantics.}
By emphasizing critical details and filtering out noise from lengthy inputs, our \ourname\ prioritizes relevant information—a crucial factor for success in open-domain QA tasks.
In contrast, \mytext\ degrades when more passages are provided to the encoder, as additional passages introduce more noise and redundancy, making it harder to extract relevant answers.
We also investigate the effectiveness of using only the fast path of \our\ (\ie, vision encoder) on the Open-domain QA task, which requires the model to generate accurate answers based on given relevant passages. 
Specifically, we feed only the top-10 relevant passages to the visual encoder (\ie, $k_\text{e}=10$), without providing any passages to the LLM directly (\ie, setting 
$k_\text{d}=0$). 
Surprisingly, this configuration yields performance on par with TinyLLaMA, despite the latter processing all 10 passages with a much heavier LLM. 
This demonstrates that the fast path of our method can distill and preserve the critical information from long contexts, providing a compact yet effective representation.

\subsection{Ablation Study}
We explore the effects of \ding{182} the masking strategy employed in PVE, \ding{183} the length of text provided to the encoder during training, \ding{184} the number of compressed tokens in each image (\ie, $N$ in \S\ref{sec::token_reduction}), and
\ding{185} extension to other LLM.
In open-domain QA tasks, the model is fed 10 passages for LLM and 5 for encoder. 
ICL tasks use a fixed setup of 18 demonstrations for encoder and 2 for LLM.
% (More details in Appendix~\ref{secg}.)

\begin{figure}[!t]
\begin{minipage}[t]{\textwidth}
\begin{minipage}{0.50\textwidth}

    \centering
      \captionsetup{font=small}
  \makeatletter\def\@captype{table}\makeatother\caption{\small{The effect of visual token count for each image.}}
  \label{tab:ab_token_num}
  \vspace{-7pt} 
  \resizebox{\textwidth}{!}{
       \begin{tabular}{cccccc}
    \toprule
        \textbf{Tokens} & \multicolumn{2}{c}{\textbf{ICL}} & \multicolumn{3}{c}{\textbf{Open-domain QA}}
        \\
        \cmidrule(lr){2-3} \cmidrule(lr){4-6} 
       \textbf{Per Image}  &  {TREC} & {MR} & {TriviaQA} &{NQ} & {PopQA} \\
    \midrule
            
        32 & 32.7 & 78.8 &   19.38& 7.85 &8.16 \\
        64 &  \textbf{36.5} & 79.2 &  \textbf{25.20} & \textbf{8.71} & \textbf{11.44}     \\
        96 & 32.0 & 79.7 &    14.57&	7.19	& 4.88 \\
        128 & 32.9 & \textbf{87.0} &  20.01 &7.77 &8.51   \\
    \bottomrule
    \end{tabular}
    } 
  \vspace{4pt}
  \centering
  \captionsetup{font=small}
  \makeatletter\def\@captype{table}\makeatother\caption{\small{{The effect of different masking strategies in PVE.} FM is frequency-based masking strategy, RM is random masking. The masking ratio is set to 50\%.}}
  \label{tab:ab_strategy}
    \vspace{-7pt} 
  \resizebox{\textwidth}{!}{
  \renewcommand\arraystretch{1.05}
        \begin{tabular}{cccccccc}
    \toprule
        &   & \multicolumn{2}{c}{\textbf{ICL}} & \multicolumn{3}{c}{\textbf{Open-domain QA}}
        \\
        \cmidrule(lr){3-4} \cmidrule(lr){5-7} 
        \multicolumn{1}{c}{\multirow{-2}{*}{\hspace{0.5mm}\textbf{RM}   \hspace{0.5mm}}}    & \multicolumn{1}{c}{\multirow{-2}{*}{\hspace{0.5mm}\textbf{FM}   \hspace{0.5mm}}}  &  {NLUS} & {NLUI} & {TriviaQA} &{NQ} & {PopQA} \\
        \midrule
            
            &   & 9.9 & 26.4 & 17.14 & 6.51 & 5.72	 \\
           \cmark  &  &  8.3 & 30.2 & 24.88 & 8.35 & 10.19	  \\
          & \cmark & \bf15.6 & \textbf{40.6} &  \textbf{25.20} & \textbf{8.71} & \textbf{11.44} \\

            \bottomrule
        \end{tabular}
    }
    \vspace{3pt}
\end{minipage}
\hspace*{2pt}
\begin{minipage}{0.50\textwidth}
  \centering
      \captionsetup{font=small}
  \makeatletter\def\@captype{table}\makeatother\caption{\small{{Ablation on the length of text inputs (in tokens).} 
  }}
  \label{tab:ab_encoder_len}
    \vspace{-7pt} 
  \resizebox{\textwidth}{!}{
  \renewcommand\arraystretch{1.05}
        \begin{tabular}{cccccc}
    \toprule
        \textbf{Encoder} & \multicolumn{2}{c}{\textbf{ICL}} & \multicolumn{3}{c}{\textbf{Open-domain QA}}
        \\
        \cmidrule(lr){2-3} \cmidrule(lr){4-6} 
       \textbf{Input Length}  &  {SST5} & {MR} & {TriviaQA} &{NQ} & {PopQA} \\
        \midrule

        1024 & 39.6 & 85.9 &   19.77&	6.47 &	6.63    \\
        2048 &  39.3& 73.8 &   22.85	&8.08	&9.77    \\
        4096  & \textbf{42.7} & 79.2 &   25.20 &	8.71 & 11.44  \\
        6144 & 37.8 & \textbf{90.5} &   \textbf{27.52}	& \textbf{9.24}&	\textbf{13.49}   \\

            \bottomrule
        \end{tabular}
    }
 \vspace{4pt}
  \centering
  \captionsetup{font=small}
  \makeatletter\def\@captype{table}\makeatother\caption{\small{\ourname\ effectively reduces PPL on LCM and improves accuracy on ICL by processing additional context.} $\ddagger$ means our implementation on Mistral 7B~\cite{jiang2023mistral}.}
\label{tab:extension_llm}
    \vspace{-7pt} 
  \resizebox{\textwidth}{!}{
  \renewcommand\arraystretch{1.15}
        \begin{tabular}{ccccc}
    \toprule
           & \multicolumn{2}{c}{\textbf{LCM}} & \multicolumn{2}{c}{\textbf{ICL}}
        \\
        \cmidrule(lr){2-3} \cmidrule(lr){4-5} 
       \multicolumn{1}{c}{\multirow{-2}{*}{\hspace{0.5mm}\textbf{Method}   \hspace{0.5mm}}}  &  {Arxiv} & {Book} & {SST2}  & {DBP} \\
    \midrule
         
        Mistral & 2.93 & 12.82 &   89.1 & 93.6 \\
        CEPE$^{\ddagger}$ & 2.83\greenp{0.10$\downarrow$} & 12.64\greenp{0.18$\downarrow$} & 90.8\greenp{1.7$\uparrow$}  & 94.2\greenp{0.6$\uparrow$}      \\
        
        \ourname$^{\ddagger}$ & \bf2.82\greenp{0.11$\downarrow$} & \bf12.61\greenp{0.21$\downarrow$} & \bf 92.8\greenp{3.7$\uparrow$}	& \bf 95.3\greenp{1.7$\uparrow$} \\

    \bottomrule
    \end{tabular}
    }
    \vspace{3pt}
\end{minipage}
\end{minipage}
\vspace{-13pt}

\end{figure}

% we conduct a series of ablation studies to examine how different training configurations affect downstream task performance. Specifically,
% \alex{In-context Text Length}
% \alex{Number of tokens in each image}
% \input{tabs/ab_strategy}
% \input{tabs/ab_image_token_nums}
% \input{tabs/ab_encoder_len}
  
\noindent\textbf{Number of Tokens in Each Image.}
The Resampler transforms image features into a fixed number of visual tokens. Table~\ref{tab:ab_token_num} analyzes the impact of visual token count for each image. 
Increasing the number of visual tokens reduces compression ratio, but as shown, \textit{\textbf{a lower compression ratio does not always yield better results}}. 
With 64 visual tokens, the model performs best on 4 out of 5 datasets, whereas 128 tokens only perform best on MR dataset. 
This discrepancy could be attributed to the trade-off between the amount of information preserved and the noise introduced during compression. 
Fewer tokens risk losing critical details, while more tokens may retain irrelevant information, which can hinder generalization.  
These findings emphasize that achieving a balance between compression and information retention is crucial for optimal performance across different datasets.

\noindent\textbf{Masking Strategy in PVE (\S\ref{sec::FCL}).} 
Long texts often contain significant redundancy. 
To address this, we integrate a Frequency-based Masking (FM) strategy into PVE, improving the information density of text token embeddings. Table~\ref{tab:ab_strategy} compares the performance of \ourname\ with FM, w/o FM, and with random masking.  
Excluding FM causes a notable decline in ICL and open-domain QA performance. 
This highlights the critical role of information-dense text token embeddings in guiding the visual encoder to capture more semantically meaningful and discriminative features.
% , ultimately boosting task performance. 

% \input{figs/random_mask}
\begin{figure}[!t]
    \centering
    \includegraphics[width=1.0\textwidth]{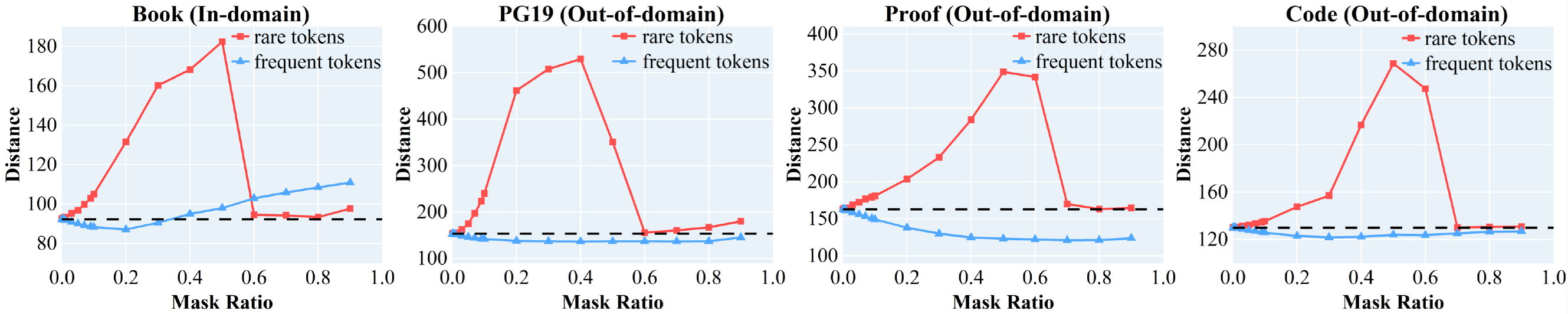} 
    
\vspace{-5pt}
    \caption{\small{\textbf{Effect of frequent \textit{vs.} rare tokens on semantic integrity.} Masking rare tokens (red) significantly disrupts semantic representation, increasing text-visual embedding distance, while masking frequent tokens (blue) has minimal impact, demonstrating that rare tokens are critical for preserving semantic meaning. }}
    \label{fig:fm}
\vspace{-15pt}
\end{figure}

% ($\hat{\bm{F}}_\text{v}$ and $\hat{\bm{F}}_\text{a}$)

\noindent\textbf{In-context Text Length.} Table~\ref{tab:ab_encoder_len} investigates the impact of in-context text length (measured in text tokens) provided to the visual encoder during training. 
Our model, trained with a longer encoder input length, generally yields higher EM scores on open-domain QA tasks. 
This may be because exposure to more lengthy texts during training helps our model better extract key information from extensive contexts, which is crucial for open-domain QA. 
Table~\ref{tab:ab_encoder_len} shows that longer training text inputs often boost ICL task accuracy. 
For instance, the best result on SST5 (42.7) was achieved with an encoder input length of 4096 text tokens, while the highest accuracy on MR (90.5) was obtained with the longest input length of 6144 tokens. 
Interestingly, we observed that training with an input length of 1024 tokens performed comparably to 2048 tokens,  possibly because the total demo length for the encoder was close to 1024 tokens.

\noindent\textbf{Extension to Other LLM.}
Mixture-of-expert models effectively scale model capacity while saving resources.
To prove the generality of \ourname, we also apply \ourname\ to Mistral 7B~\cite{jiang2023mistral}. 
In LCM task, \ourname$^{\ddagger}$ and CEPE$^{\ddagger}$ process 4096 tokens,  compared to the 2048 tokens processed by Mistral. 
For ICL, \ourname$^{\ddagger}$ and CEPE$^{\ddagger}$ use 20 demonstrations, while Mistral uses only 2. 
As shown in Table~\ref{tab:extension_llm}, \ourname$^{\ddagger}$ demonstrates superior performance over CEPE$^{\ddagger}$, by effectively leveraging additional context.
% To further validate the scalability of our method, results on larger-scale LLMs are included in Appendix~\ref{secg}.

% llama2 sdpa 4bit deepspeed zero stage2 bfloat16 length 512+4096

% \alex{\subsection{Some Visualizations}}
% \subsection{Token Redundancy Exploration}
% % \alex{Give More Specific Title, like token redundancy exploration}
% We study the semantic contribution of frequent tokens by selectively masking the top 50\% most frequent tokens (based on training set statistics).
% Figure~\ref{fig:mask_exam} illustrates key nouns are fully preserved. 
% The masked tokens mainly include function words(\eg, ``the", ``of"), primarily serving grammatical roles rather than conveying core semantic meaning. 
% This proves FM strategy can
% \textbf{preserve critical semantic information while filtering out less relevant noise, thereby enhancing the ability of the model to focus on meaningful content}.

\begin{figure}[t]
  \centering
  \begin{minipage}[t]{0.48\textwidth}
    \centering
    \includegraphics[width=\textwidth]{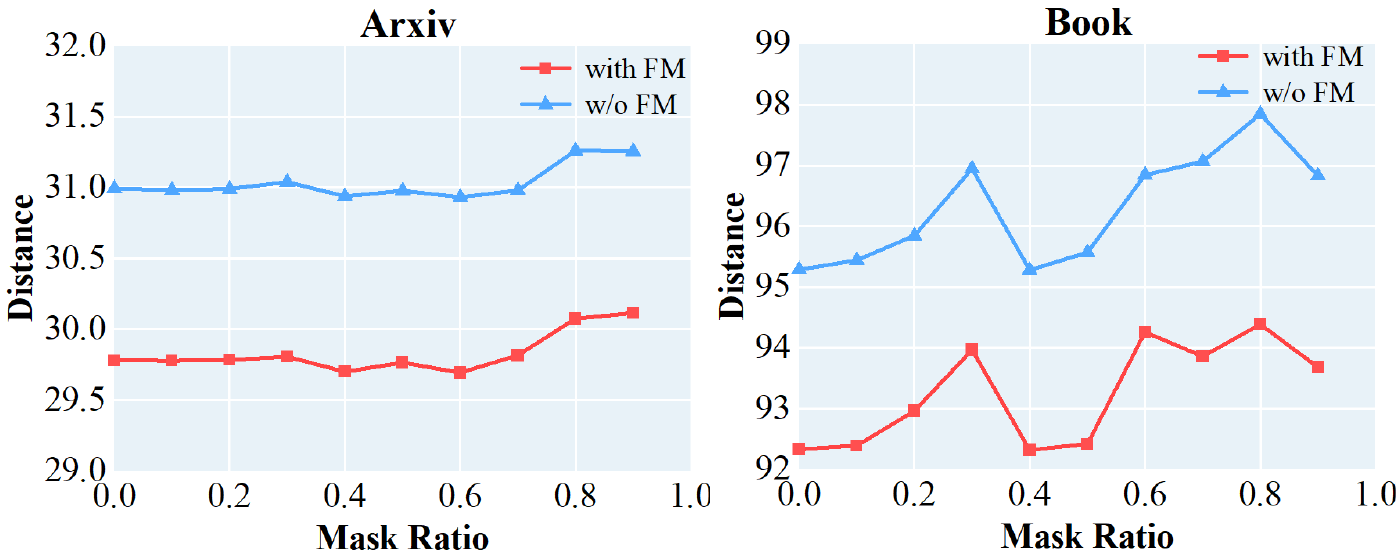}
    \vspace{-15pt}
    \caption{\small{\textbf{Impact of Frequency-based Masking on Text-Visual Semantic Distance.} Compared to model without FM, \ourname\ with FM consistently reduces the semantic distance between text token embeddings and visual features, across all masking ratios.}}
    \label{fig:random_mask}
  \end{minipage}%
  \hfill
  \begin{minipage}[t]{0.48\textwidth}
    \centering
    \includegraphics[width=\textwidth]{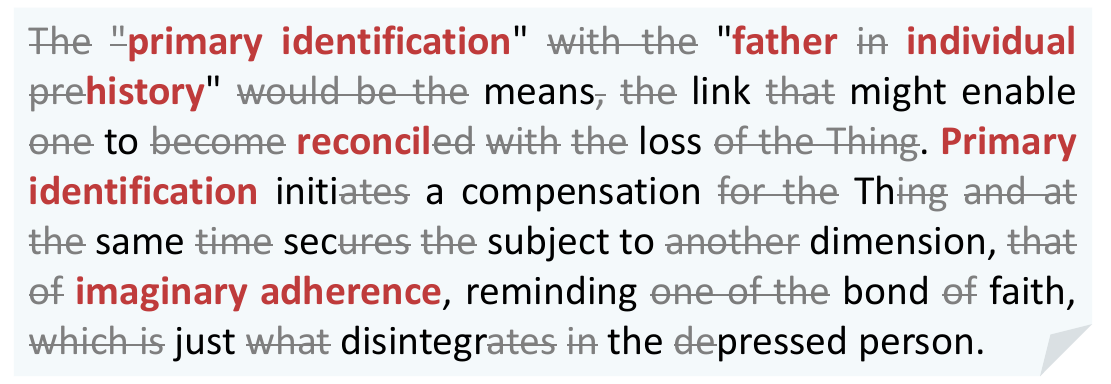}
    \vspace{-15pt}
    \caption{\small{\textbf{Visualization of Masking Frequent Tokens.} 
Though we mask the top 50\% most \textcolor[rgb]{0.5,0.5,0.5}{\sout{frequent tokens}} (based on training statistics), \textcolor[rgb]{0.75,0.23,0.23}{\textbf{key nouns}} are completely preserved, proving that frequent tokens contribute less to semantic meaning.}}
    \label{fig:mask_exam}
  \end{minipage}
\vspace{-15pt}
\end{figure}

\section{Discussion}
\label{sec::discussion}
% \subsection{rendering strategy in reference stage}
% ICL task render方式：1. 均匀分布 指定图像数量 2 3 2. 每个图片一个demo 3. 每个图片一个demo并且添加很多的空格，相当于换行操作 这也实现了不同的压缩比例。。
% on global semantic alignment between text tokens and visual tokens.

% add to the appendix?
% In our experiments, random masking ratios (0.0 to 0.9) were applied to text token embeddings. For each sample, we computed the cosine distance between the pooled text token embeddings and visual features. The distance was then calculated as the sum of these distances across all test samples in the Arxiv and Book datasets.

\noindent\textbf{Exploring Token Frequency as a Proxy for Semantic Importance.}
% Visualizing the Contribution of Different Tokens to Semantics.
To assess the impact of rare versus frequent text tokens on global semantics across in-domain$_{\!}$ (Book~\cite{weberredpajama}) $_{\!}$and out-of-domain datasets$_{\!}$ (PG19~\cite{raecompressive}, Proof~\cite{azerbayev2023proofpile}, and Code~\cite{codep}), we first calculate importance score for each token using Eq.$_{\!}$~\ref{eq:important_score} based on training-set token frequency statistics. Two masking operations are then applied to the text token embeddings: $_{\!}$\ding{182} Masking tokens with high importance scores (red line in Figure$_{\!}$~\ref{fig:fm}). $_{\!}$\ding{183} Masking tokens with low importance scores (blue line in Figure$_{\!}$~\ref{fig:fm}). $_{\!}$The distance between masked text embeddings and visual features is computed.

At low masking ratios (0.0 to 0.4), masking rare tokens causes a sharp increase in distance, while masking frequent tokens has minimal impact and even reduces distance. This suggests that rare tokens carry more critical semantic information, and their removal disrupts the alignment between text and visual tokens. In contrast, frequent tokens may contain more redundant or less informative content, so masking them has little impact or even improves the alignment by reducing noise. 
These findings support the hypothesis that \textit{token frequency is a reasonable indicator of semantic importance, with rare tokens playing a more pivotal role in preserving semantic integrity}.
% These findings highlight the importance of rare tokens in preserving semantic integrity and demonstrate the potential benefits of filtering out frequent, less informative tokens.  
% \input{figs/mask_demo}

\noindent\textbf{The Effect of Frequency-based Masking on Text-Visual Semantic Gap.}
\ourname\ employs Frequency-based Masking (FM) within the Probability-informed Visual Enhancement to improve the semantic richness of text token embeddings.
Figure~\ref{fig:random_mask} presents the impact of FM on the semantic alignment between text token embeddings and visual features extracted by the Perceiver Resampler. 
We experimented with random masking ratios (0.0 to 0.9) on text token embeddings and calculated the sum of cosine distances across all test samples in the Arxiv and Book datasets~\cite{weberredpajama}. 
Across all ratios, \textit{\ourname\ with FM consistently exhibits smaller semantic distances than \ourname\ without FM, highlighting FM effectively enhances semantic coherence}.

\noindent\textbf{Token Redundancy Visualization.}
We study the semantic contribution of frequent tokens by selectively masking the top 50\% most frequent tokens (based on training set statistics).
Figure~\ref{fig:mask_exam} illustrates key nouns are fully preserved. 
The masked tokens mainly include function words (\eg, ``the'', ``of''), primarily serving grammatical roles rather than conveying core semantic meaning. 
This proves FM strategy can
\textbf{preserve critical semantic information while filtering out less relevant noise, thereby enhancing the ability of the model to focus on meaningful content}.

\section{Conclusion}
In-context learning with longer input sequences remains a prominent yet challenging topic in large language models (LLMs). 
In this work, we introduce a fully novel perspective to address this challenge by leveraging much lightweight visual encoder.
To support longer input sequences in LLMs, we present \ourname, a vision-centric token expansion method built upon a visual encoder framework. 
Our analysis further reveals there exists significant redundancy in text tokens, further validating the effectiveness and efficiency of our vision-encoder-based approach.
With these advancements, \ourname\ surpasses text-encoder-based token compression counterparts in both performance and efficiency. 
In future work, we plan to evaluate \ourname\ across a broader range of downstream tasks and conduct a deeper investigation into text token redundancy.

{\small
\bibliographystyle{unsrt}
\bibliography{neurips_2025}
}

% \begin{ack}
% Use unnumbered first level headings for the acknowledgments. All acknowledgments
% go at the end of the paper before the list of references. Moreover, you are required to declare
% funding (financial activities supporting the submitted work) and competing interests (related financial activities outside the submitted work).
% More information about this disclosure can be found at: \url{https://neurips.cc/Conferences/2025/PaperInformation/FundingDisclosure}.

% Do {\bf not} include this section in the anonymized submission, only in the final paper. You can use the \texttt{ack} environment provided in the style file to automatically hide this section in the anonymized submission.
% \end{ack}

% \newpage
% \input{sec/checklist}

%%%%%%%%%%%%%%%%%%%%%%%%%%%%%%%%%%%%%%%%%%%%%%%%%%%%%%%%%%%%
% \newpage
% \appendix
% \input{sec/appendix}

% \section{Technical Appendices and Supplementary Material}
% Technical appendices with additional results, figures, graphs and proofs may be submitted with the paper submission before the full submission deadline (see above), or as a separate PDF in the ZIP file below before the supplementary material deadline. There is no page limit for the technical appendices.

%%%%%%%%%%%%%%%%%%%%%%%%%%%%%%%%%%%%%%%%%%%%%%%%%%%%%%%%%%%%

\end{document}